\documentclass[a4paper]{article}
\pdfoutput=1
\usepackage[utf8]{inputenc}
\usepackage{lmodern}

\usepackage[english]{babel}
\usepackage{csquotes}

\usepackage[utf8]{inputenc}
\usepackage[english]{babel}
\usepackage{graphicx}
\graphicspath{ {./images/} }
\usepackage{xcolor}
\usepackage{textcomp}
\usepackage[T1]{fontenc}
\usepackage{geometry}
\usepackage{setspace}
\usepackage{graphicx} 
\usepackage{subfig} 
\usepackage{caption}
\usepackage{lipsum}
\usepackage{gensymb}
\graphicspath{ {./images/} }

\usepackage[nottoc]{tocbibind} 

\usepackage{hyperref}
\hypersetup{
  pdfinfo={
    Title={Mapping Areas using Computer Vision Algorithms and Drones}
  }
}
\usepackage{pdfpages}

\title{\textbf{Mapping Areas using Computer Vision Algorithms and Drones}}
\author{
   Bashar Alhafni\and
   Saulo Fernando Guedes\and
   Lays Cavalcante Ribeiro\and
   Juhyun Park\and
   Jeongkyu Lee\and\\
   University of Bridgeport. Bridgeport, CT, 06606. United States of America\\
   \{balhafni, sguedesd, lcavalca, juhpark\}@my.bridgeport.edu, jelee@bridgeport.edu 
}
\date{}

\begin{document}

\maketitle

\section*{Abstract}
The goal of this paper is to implement a system, titled as Drone Map Creator (DMC) using Computer Vision techniques. DMC can process visual information from an HD camera in a drone and automatically create a map by stitching together visual information captured by a drone. The proposed approach employs the Speeded up robust features (SURF) method to detect the key points for each image frame; then the corresponding points between the frames are identified by maximizing the determinant of a Hessian matrix. Finally, two images are stitched together by using the identified points. Our results show that despite some limitations from the external environment, we could have successfully stitched images together along video sequences. 
\section*{Keywords}
Image Stitching, Drone, OpenCV, BoofCV, JAVA
\section{Introduction}
Drones are getting popular in a wide range of research in recent years. Many companies have developed their own drone technology such as Google, Facebook, and Amazon. In addition, drones are used in “Drone Journalism” so they can obtain videos of areas that are hard to reach. Besides being used for commercial reasons, drones can be used for military, security, exploration, and surveillance. A quadcopter is one kind of drone that has four rotors. Two spin clockwise, and the other two spin counter-clockwise. They work in tandem to balance the drone. Additionally, a drone can fly autonomously based on a pre-entered program without piloting. Since most of drones carry video cameras with considerably high resolution, the use of them for image processing and computer vision techniques has become a very interesting topic. Applications using drones and cameras have been developed to solve many problems and make people’s lives easier. One of challenges of such drone applications is that a drone cannot work properly when a GPS signal or a pre-captured satellite map is not available. This research will address the limitation by proposing a drone map creator that utilizes computer vision techniques for video streams captured by a drone.\\

In our research project, our environment was the university parking lot, and we used image processing techniques to reach our goal. In this paper, we will show how we get the images by remotely controlling the AR-Drone to feed the image stitching algorithm, which will generate a big image by blending these images together.

\section{Background}

The aim of our research project is to come up with an image processing algorithm that will let us cover the University of Bridgeport parking lot, and this can be done by taking successive images and then combining these images into one large image. To have images with overlapped areas so that they can be combined, we needed a stable device with a good quality camera, which will capture the images in a certain order. That’s why we chose the AR-Drone 2.0.\\

The AR-Drone\cite{drone} operator can directly set its yaw, pitch, roll, and vertical speed, and the control board adjusts the motor speeds to stabilize the drone at the required pose. The first camera with approximately 75\textdegree\ × 60\textdegree\ field of view is aimed forward and provides a 640 × 480 pixel color image. The second one is mounted on the bottom and provides color image with 176 × 144 pixels and its field of view is approximately 45\textdegree\ × 35\textdegree. Therefore, the front camera of the drone will provide us with good quality stabilized overlapped images which can be combined into one image using the algorithm that we developed. Along with the AR-Drone 2.0, we also used existed free software libraries that we found helpful to develop our algorithm. The libraries are OpenCV and BoofCV.\\

OpenCV\cite{opencv} (Open Source Computer Vision Library) is an open source computer vision and machine learning software library. The library has many optimized algorithms, which can be used in many areas such as face recognition, identifying objects, tracking moving objects, and stitching images together to produce a high resolution image of an entire scene. OpenCV also leans mostly towards real-time vision applications and it has many available interfaces (C++, Python, Java, etc.). According to all of these properties, we used the Java interface of OpenCV, which is a primary component of our algorithm.\\

In addition to OpenCV and all the image processing features it has, BoofCV\cite{boofcv} has also demonstrated very high-level image processing capabilities. BoofCV is a Java library for real-time computer vision and robotics applications. It includes low-level image processing routines, feature tracking, and geometric computer vision. BoofCV is organized into several packages: image processing, geometric vision, calibration, recognition, and visualization. The library has also an example for an image stitching algorithm, which is the main goal of our research project.\\

Image stitching refers to combining two or more overlapping images together into a single large image. When stitching images together, the goal is to find a 2D geometric transform which minimizes the error (difference in appearance) in overlapping regions. There are many ways to do this. BoofCV uses an example where point image features are found, associated, and then a 2D transform is found robustly using the associated features.\newpage

Another way to achieve image stitching is by the Scale Invariant Feature Transform (SIFT) algorithm. According to Andrea Vedaldi\cite{VLFeat.org}, a SIFT feature is a selected image region (also called keypoint) with an associated descriptor. Additionally, there’s another image stitching technique on which our algorithm is based on. This technique is called Speeded up robust features (SURF). The SURF and SIFT techniques are very similar but they have a few minor differences in some details. The main concept of our image stitching algorithm is based on the OpenCV library as well as the SURF algorithm, which gave us the results that we almost were aiming for.

\section{Objective and Approach}

Our goal in this project is to use the drone’s camera to take successive pictures from the university parking lot and blend those images together to build a new image, which will cover the entire parking lot. To do that, we implemented two approaches. First, we used the BoofCV library that has its own image stitching algorithm and features. Second, we developed our own algorithm based on the OpenCV library and the SURF algorithm.\\

The images were taken during certain periods of time so that the drone captures overlapped images. This way, we can identify and adjust these areas to create one large image. To perform our code in the field test, we divided our environment like the image below. We had the parking lot divided in a 4x4 matrix so that we could organize and better control the drone’s movement.  

\begin{figure}[ht]
\centering
  \includegraphics{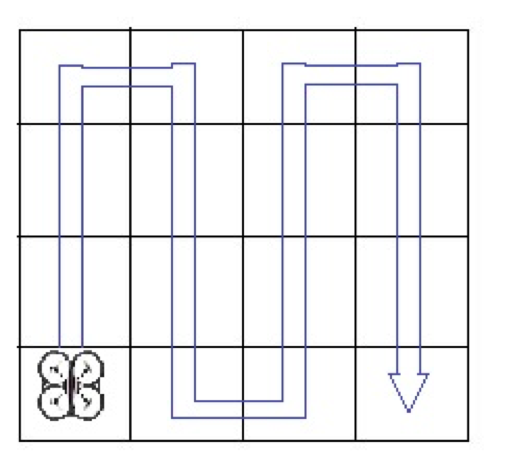}
  \caption{Drone's movement grid}\label{fig:figures}
\end{figure}

The first step in our approach is to take an image at our starting point. Then, the drone will move forward taking pictures to cover the first column of our matrix. After completing the first column, the drone will move to realign with the next column, but now taking pictures in the opposite direction. We repeated these steps until the matrix was fully covered.  Each column is composed by four smaller images that were stitched together, then we stitched the columns together after receiving appropriate modifications, such as rotating the images.  To perform the stitching algorithm, we had two cases in our environment: first, stitching images from bottom to top. Second, stitching images from side to side.

\subsection{First algorithm: BoofCV}

The BoofCV library has its own image stitching algorithm. This algorithm, finds some image points features and then a 2D transform is found robustly using the associated features. This image stitching algorithm can be summarized as follows: first, detect and describe point features, which is the discovery of the best key points of each image. Then associate features together, only with the common key points between the images. After that, apply a robust fitting to find a transform that will make the necessary changes in the images, such as rotations. The final step is to render the combined images to generate the final image.  
\begin{figure}[ht]
\centering
  \includegraphics[scale=0.5]{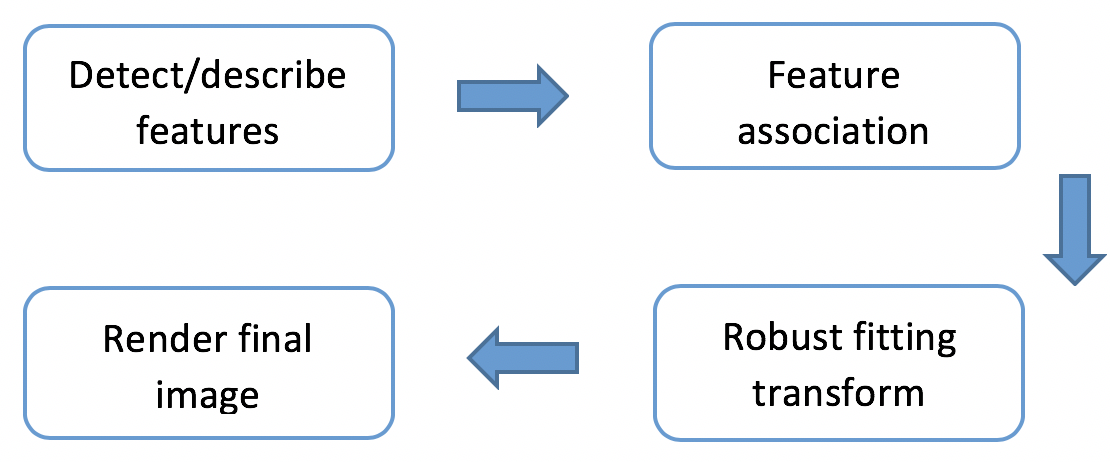}
  \caption{BoofCV algorithm steps}\label{fig:figures}
\end{figure}

Once BoofCV renders two images together, it generates a single image with a black background to complete the square referring to the image file itself. After that, when we try to stitch this result with a third image, we will not get the desired result. This happens because the black background will be considered as a part of the image, so the algorithm will try to find a correspondence between the background and some features of the third image and will not find the correct correlation between the images.\\

Even if we use some techniques to cut off the background, the result will still present distortions. This will happen because of the translations that BoofCV does. These translations change the rectangular shape of the image, so there will always be some black background. Thus, the result of stitching more than two images using BoofCV will be a very distorted image. To avoid this problem, we decided to implement our own algorithm, using OpenCV.

\subsection{Stitching algorithm using OpenCV}
Our stitching algorithm was developed based on the OpenCV library and mostly prepared using the image processing technique SURF. To perform the main feature of our algorithm, the key point detection using SURF is our first and main step.\\

According to Herbert Bay\cite{surf}, the SURF algorithm is a local feature detector and descriptor that can be used for tasks such as object recognition or registration or classification or 3D reconstruction. It is partly inspired by the scale-invariant feature transform (SIFT) descriptor. The SURF algorithm is based on the same principles and steps as the SIFT, but details in each step are different. The algorithm has three main parts: interest point detection, local neighborhood description and matching. SURF uses a blob detector based on the Hessian matrix to find points of interest. The determinant of the Hessian matrix is used as a measure of local change around the point and points are chosen where this determinant is maximal. \\

\begin{figure}[ht]
\centering
  \includegraphics{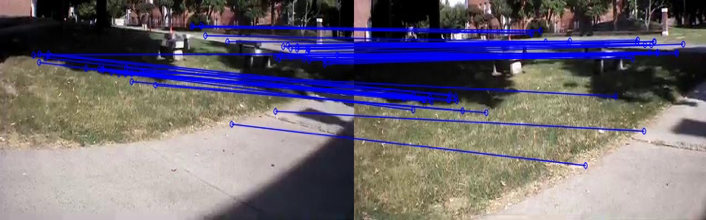}
  \caption{Key points and feature detection from two consecutive images}\label{fig:figures}
\end{figure}

After identifying the key points and matching them between the images, we must identify the common area between these images and crop it. \\

\begin{figure}[ht]
\centering
\subfloat[Intersection in the first input]{\label{fig:mdleft}{\includegraphics[width=0.45\textwidth]{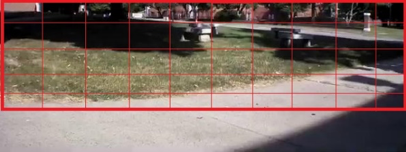}}}\hfill
\subfloat[Intersection in the second input]{\label{fig:mdright}{\includegraphics[width=0.45\textwidth]{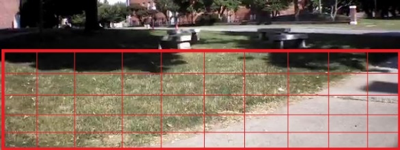}}}\hfill
\caption{Intersections in the input images}
\label{fig:subfigures}
\end{figure}

\begin{figure}[ht]
\centering
  \includegraphics{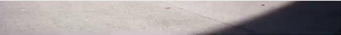}
  \caption{new first image without the intersection}\label{fig:figures}
\end{figure}
\begin{figure}[ht]
\centering
  \includegraphics{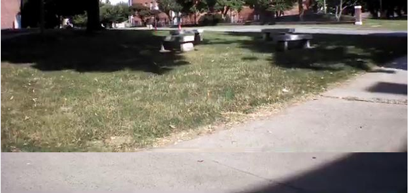}
  \caption{Final image after the stitch algorithm}\label{fig:figures}
\end{figure}
Now, the first image has only the part that is not included in the second image. We always keep the last image for performance and level details. The next step is simply to blend those two images together; the first in the bottom and the second in the top. The result of this rendering will be the first input to the algorithm followed by the pictures taken when the drone moves forward.

\newpage
\section{Results}
We took three pictures (Figures 7(a), 7(b) and 7(c)) with the front camera of the drone and cut them to feed the algorithm. We were able to get two satisfactory results, even though they have very visible differences, as shown in Figure 7(a) and Figure 7(b).
\begin{figure}[ht]
\centering
\subfloat[Image 1]{\label{fig:mdleft}{\includegraphics[width=0.32\textwidth]{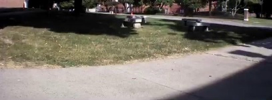}}}\hfill
\subfloat[Image 2]{\label{fig:mdright}{\includegraphics[width=0.32\textwidth]{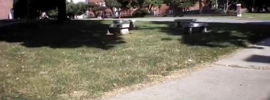}}}\hfill
\subfloat[Image 3]{\label{fig:mdright}{\includegraphics[width=0.32\textwidth]{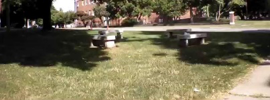}}}
\caption{Results}
\label{fig:subfigures}
\end{figure}

The stitching algorithm provided by the BoofCV library does not have a complete rectangular image because it rotates the images for a better binding, which means that the image will have gaps that will be filled by a black background, as shown in Figure 8(a). On the other hand, the algorithm implemented by us simply binds the images together without any gaps and black backgrounds, as shown in Figure 8(b).

\begin{figure}[ht]
\centering
\subfloat[BoofCV Result]{\label{fig:mdleft}{\includegraphics[width=0.45\textwidth]{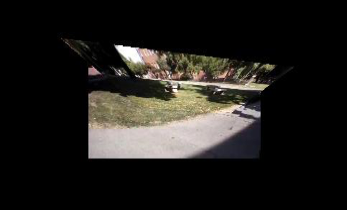}}}\hfill
\subfloat[Stitched Images]{\label{fig:mdright}{\includegraphics[width=0.45\textwidth]{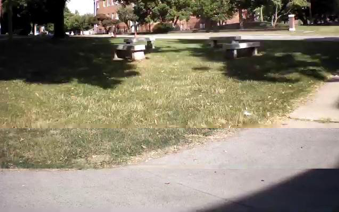}}}\hfill
\caption{BoofCV vs our results}
\label{fig:subfigures}
\end{figure}

It is clear that the BoofCV algorithm binds two images with more precision than our implementation, and we barely see the transition between these two images, but it will present a lot of distortions as soon as we attempt to stitch a higher number of images. This distortion makes it difficult to bind more images because increasing the amount of images will increase the amount of distortions, which will decrease the quality the results. However, our algorithm will not present distortions, but it will have clear lines of transitions between the bound images. Since some details of the images can get lost during this process, this will also affect the quality of the results. Therefore, classifying which algorithm works better for this problem is a difficult decision.\\

\section{Conclusion}
After running some examples, we decided that, the best approach for creating a bigger image from other small images is the algorithm developed by us. Even though BoofCV can fit two images with better accuracy, the drone mapping will have a large number of images to stitch, and BoofCV's results would be impracticable because of the amount of distortion it presents. Thus, the algorithm developed by us is a good candidate for the proposed task. \\

At the end of this project, we could not reach the goal of mapping the parking lot. Working with image stitching algorithms requires very high quality images, which unfortunately the AR-Drone 2.0 does not provide to us. One good possibility for completing this task is using the new version of it, or even using some different drone, with preferentially a 1080p camera. Another problem while dealing with the use of drones for this project is the low stability of the drone. Because of external environment, the pictures taken by the drone were often without a common area, so the algorithm did not perform as expected. Thus, the best drone to be used in such project is one with not only a high quality camera, but also with a good stability. With the efficient resources, a parking lot mapping can be done, and it can be used to create a very reliable vigilance system.\\

\end{document}